\documentclass{article}

\usepackage[letterpaper,top=2cm,bottom=2cm,left=3cm,right=3cm,marginparwidth=1.75cm]{geometry}
\usepackage{packages/my_style,times}
\usepackage{fancyhdr}
\usepackage{cite}
\usepackage{graphicx}
\reportfinal

\title{\noindent Architectural Implications of Embedding Dimension during GCN on CPU and GPU}
\author{Matthew Adiletta, David Brooks, Gu-Yeon Wei \\
Harvard University}

\begin{document}
\maketitle

\newcommand{\matt}[1]{{{\footnotesize\color{orange}[matt: #1]}}}

\newcounter{takeaways}
\setcounter{takeaways}{1}
\newcommand{\takeaway}[1]{\smallskip\noindent\fbox{\parbox{.965\linewidth}{\small\textbf{Key Takeaway \thetakeaways: }\emph{#1}\addtocounter{takeaways}{1}}}\smallskip}

\newcommand{\highlight}[1]{\smallskip\noindent\fbox{\parbox{.965\linewidth}{\emph{#1}}}\smallskip}
\newcommand{\linehighlight}[1]{\emph{\textbf{#1}}}

\begin{abstract}

Graph Neural Networks (GNNs) are a class of neural networks designed to extract information from the graphical structure of data. Graph Convolutional Networks (GCNs) are a widely used type of GNN for  transductive graph learning problems which apply convolution to learn information from graphs. GCN is a challenging algorithm from an architecture perspective due to inherent sparsity, low data reuse, and massive memory capacity requirements. Traditional neural algorithms exploit the high compute capacity of GPUs to achieve high performance for both inference and training. The architectural decision to use a GPU for GCN inference is a question explored in this work. GCN on both CPU and GPU was characterized in order to better understand the implications of graph size, embedding dimension, and sampling on performance. 
\end{abstract}

% transductive: training set to test set
% inductive: general rules which can be broadly applied
\section{Introduction} 

A Graph Convolutional Network (GCN) is a type of Graph Neural Network (GNN) which applies convolution to graphically structured data \cite{defferrard_convolutional_2017}. A graph convolution layer can be disaggregated into two steps; neighborhood aggregation and update. During neighborhood aggregation, the sparse graphical structure of the data is exploited and each vertex aggregates the embedding vectors of its neighbors. During update, a convolution kernel is applied to transform the embeddings, similarly to a conventional Convolutional Neural Network (CNN). After the graph convolution layer, a non-linear activation function is applied, resulting in the input to the next convolution layer.  

Both CPUs and GPUs have been widely used for training and inference of GCNs \cite{gong_graphite_2022, md_distgnn_2021, rahman_fusedmm_2021, huang_ge-spmm_2020, selvitopi_distributed-memory_2021, zhang_architectural_2020, yan_characterizing_2020, baruah_gnnmark_2021, zhang_pcgraph_2021, wang_empirical_2021}. CPUs offer terabytes of memory capacity and a scalable computing platform while GPUs offer high memory bandwidth, latency tolerance, and massive compute parallelism. One challenge with using GPUs for Graph Convolution is memory capacity. Today's largest GPUs have memory capacity on the orders of 10s of GBs while large scale GCNs have a memory footprint on the orders of 100s of GBs to terabytes - if a graph cannot fit in a GPU's memory, alternative techniques must be explored for offloading the GCN to GPU.

GCN mini-batching with vertex aware clustering is one technique used to overcome memory capacity limitations when using a GPU. Mini-batching involves creating a subgraph from a set of vertices and their N-hop  neighbors \cite{hamilton_inductive_2018}. Minibatch sampling can be accomplished using batch-wise sampling or layer-wise sampling. 

Batch-wise sampling starts by sampling all the vertices needed for a mini-batch computation. This can be accomplished by clustering vertices together and sampling vertices from a cluster, or randomly sampling vertices across the entire graph. Sampling within a cluster helps preserve data locality by increasing the reuse of embeddings; although clustering adds additional overhead which may not be tolerated in some cases. 

After a set of vertices $V_0$ are sampled, all neighboring vertices must be discovered. Using a neighborhood sampling algorithm, the set of neighbors $V_1$ are found. The computation of the 1-hop intermediate representation for vertices $V_0$ can be found using neighborhood aggregation. For the next layer, all neighbors of set $V_1$ are added to the set of vertices $V = V_0 \cup V_1 \cup V_2$. This allows the computation for 2-hop intermediate representation of vertices $V_0$. For a GCN of depth $D$, sampling must occur $D$ times leading to a set of vertices $V = V_0 \cup V_1 \cup ... \cup V_{D-1} \cup V_D$. 

One problem with this batch-wise full-neighborhood algorithm is that as the number of layers increases, the number of vertices needed for the next layer's computation increases exponentially until a saturation point where almost the full graph is being used. Layer-wise sampling attempts to solve this problem by sampling within a layer. Rather than computing an output for a vertex in a single-shot, layer-wise sampling compute intermediate representations for each vertex, requiring a larger memory capacity than batch-wise sampling. The use case for layer-wise sampling and batch-wise sampling depends heavily on the sparsity of the graph and the total graph size. This relationship is not investigated thoroughly in this work. 

Another solution to minimize exploding neighborhoods is to sample using a fixed number of neighbors per vertex during aggregation. This provides a predictable run time and fixed memory capacity. The GraphSage paper demonstrated that using fixed-size, uniform sampling could achieve high performance with as few as two samples per vertex \cite{hamilton_inductive_2018}. Other works such as in Cluster Sampling \cite{dong_global_2021} use a sampling fanout of (15, 10, 5). Fixed-size sampling changes the accuracy of the model; however, some cases show that this normalization strategy can actually improve accuracy. This work does not explore fixed-sampling, as the primary investigation is a performance characterization of the vanilla GCN algorithm. Advanced sampling and clustering techniques was left to future work. 

Many works which implement vertex-sampling techniques are for training on GPUs. The reason they may not implement sampling for inference on GPU is because sampling on CPU is a large bottleneck. During sampling, the CPU must traverse a graph and touch all vertices which are required in the GPU computation. The CPU must also construct a sub-graph, aggregate embeddings, and move the data between the CPU and GPU, which is expensive. This is true especially for small graphs. 

In a production GCN setting, the model weights may be located on the GPU, while the graph and feature vectors arrive through a query. Thus, for each inference, the entire graph and feature matrix must be offloaded. Furthermore, each embedding vector in the GCN likely has low reuse (depending on the sparsity of a graph), therefore, the FLOPs/bytes ratio of GCN is also very low. GPUs excel at problems with high FLOPs/bytes ratio, which means GPUs may not be the best solution for GCNs. Finally, for mini-batching cases, each time a mini-batch is sampled, the CPU must execute the sampling, which is slow. Then a large feature vector matrix and adjacency matrix must be offloaded to GPU. Both these operations bottleneck the GPU performance and cause severe under utilization of the GPU.

\subsection{Contributions}

The performance of GCN was evaluated for a sweep of embedding dimensions. From an application perspective, the embedding dimension is a hyperparameter useful for improving model accuracy. From an architecture perspective, the embedding dimension influences the ratio of sparse to dense compute, memory capacity, memory bandwidth utilization, and end-to-end latency. The impact of embedding dimension was studied from the architecture perspective, offering the following contributions: 

\begin{enumerate}
    \item \textbf{\textit{Performance analysis of GCN on CPU}}: The execution of GCN inference on CPU is a baseline for this work. GCN inference was characterized, and the contribution of key kernels on execution time was highlighted. This is important for the research community because it reveals the importance of accelerating sparse computation while keeping memory-capacity as a first-order requirement. It also confirms that sparse computation is a performance driver for GCN across all embedding dimensions. Finally, it highlights the architectural implications of the GCN embedding dimension hyperparameter. 

    \item \textbf{\textit{Performance analysis of GCN on GPU}}: The execution time breakdown for running full-graph convolution on GPU is shown. Memory capacity restrictions on the A100-40GB GPU prevented the characterization of some large-scale graphs; however, many graphs with commonly used embedding dimensions were profiled. Bottlenecks which appear when using a GPU for full graph convolution were highlighted. Finally, GCN inference performance on GPU was compared against CPU, leading to a discussion about when GPUs are better than CPUs for GCN inference.  
    
    \item \textbf{\textit{Performance analysis of Full-Neighborhood-Sampling Graph Convolution on GPU}}: The execution time breakdown for running neighborhood sampling with full neighborhoods using both batch-wise and layer-wise techniques during graph convolution when a graph does not fit in GPU memory was shown. One concern was the unbounded nature of using full-neighborhoods; however, the batch size was carefully selected as to near-maximally utilize, but not exceed, the memory capacity of the A100. Future systems with increased memory capacity may see improvements in run time due to reduced data movement, as fewer batches would be required. 
\end{enumerate}

\subsection{Related Work}

GraphSage evolved neighborhood sampling to use a fixed-neighborhood \cite{hamilton_inductive_2018}. They noted that although the number of vertices in each layer still grows exponentially, the run-time and memory-capacity demands were more predictable than full-neighborhood algorithms.  

Other works proposed methods for optimizing sampling and offload times between CPU and GPU. LazyGCN \cite{ramezani_gcn_2020} was proposed to pipeline sampling and offload time using a two step process. First, a large batch of vertices were sampled on CPU using full-neighborhoods and offloaded to GPU. Then the GPU applied fixed-neighborhood mini-batch sampling on the batch of vertices. Trade-offs in the CPU cluster-batch size and GPU mini-batch size led to accuracy tradeoffs. 

Global Neighbor Sampling (GNS) is another technique used to reduce sampling and offload time \cite{dong_global_2021}. The GNS algorithm samples a set of vertices and stores them in the GPU memory. To select the next mini-batch, the algorithm prioritizes vertices with embedding vectors already located on GPU. The intent is to reduce data movement between CPU and GPU and speed up sampling time.

\section{Characterization Methodology}

GCN was profiled using nine benchmarks from the OGB datasets \cite{hu_open_2021}, listed in Table \ref{table:tab_ogb}. The baseline system used in this performance characterization was a dual-socket Intel(R) Xeon(R) Platinum 8380 CPU with 40 cores per socket and 512 GB of main memory using DDR4-3200. This system is one of the largest non-distributed CPU systems available. To profile GPU, an NVIDIA A100 GPU was used with 40 GB memory capacity \cite{choquette_nvidia_2021} with 32GB/sec \textbf{PCIe4.0} network fabric between host and GPU. The CPU host was a dual socket Intel(R) Xeon(R) Platinum 8380 CPU with 40 cores per socket and 512 GB of main memory using DDR4-3200.

\begin{table}[htbp] 
\vspace{-8pt}
\caption{\label{table:tab_ogb} OGB Dataset Descriptions}
\begin{center}
\begin{tabular}{ c c c c }
\hline
Name & Num Vertices & Num Edges & Density \\
\hline
ddi & 4,267 & 1,334,889 & $7.33*10^{-2}$\\
proteins & 132,534 & 39,561,252 & $2.25*10^{-3}$\\
arxiv & 169,343 & 1,166,243 & $4.06*10^{-5}$\\
collab & 235,868 & 1,285,465 & $2.31*10^{-5}$\\
ppa & 576,289 & 30,326,273 & $9.13*10^{-5}$\\
mag & 1,939,743 & 21,111,007 & $5.61*10^{-6}$\\
products & 2,449,029 & 61,859,140 & $1.03*10^{-5}$\\
citation2 & 2,927,963 & 30,561,187 & $3.56*10^{-6}$\\
papers & 111,059,956 & 1,615,685,872 & $1.31*10^{-7}$\\
\hline
\end{tabular}
\end{center}
\vspace{-8pt}
\end{table}

The methodology for profiling GCN for full-graph and  full-neighborhood-sampling approaches is detailed in this section. The characterization used the Torch-Geometric and Torch-Sparse frameworks. %\mjedit{see Appendix A for instructions on setting up the conda environment and scripts to replicate the results in this paper.} 
An embedding dimension sweep study is conducted across the hidden layers, while the input and output embedding dimensions of the graph do not change. 

\newpage
\subsection{Full-Graph GCN Characterization}

Each OGB graph was profiled on CPU and GPU for GCN using the following methodology.

\begin{figure}[t]
\centering
\includegraphics[width=.9\linewidth]{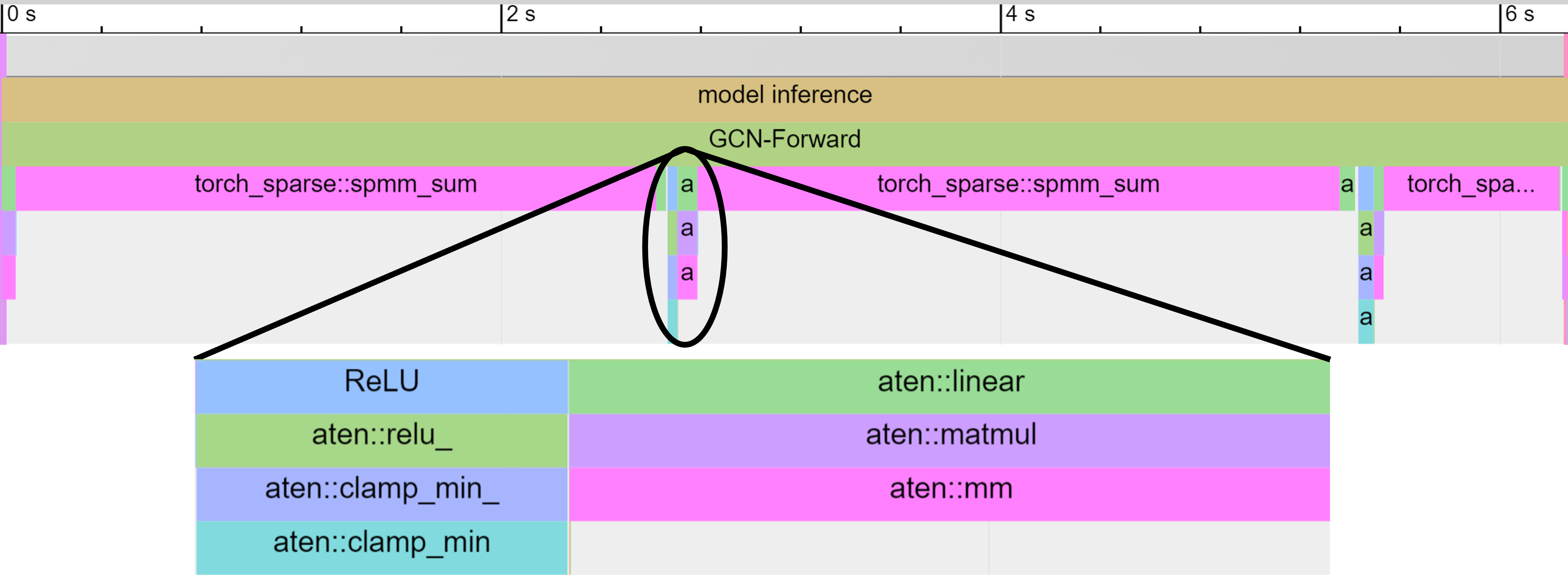}
\caption{
    CPU execution timeline for \textit{products} embedding dimension 256. This trace shows the large impact of the sparse kernel (torch\_sparse:spmm\_sum) on total execution time. The expanded section shows the activation function (ReLU) for the first GCN layer, and the dense linear matrix multiplication during the start of the second layer. 
}
\label{fig:cpu-exe-time}
\vspace{-1em}
\end{figure}

\begin{figure}[t]
\centering
\includegraphics[width=.9\linewidth]{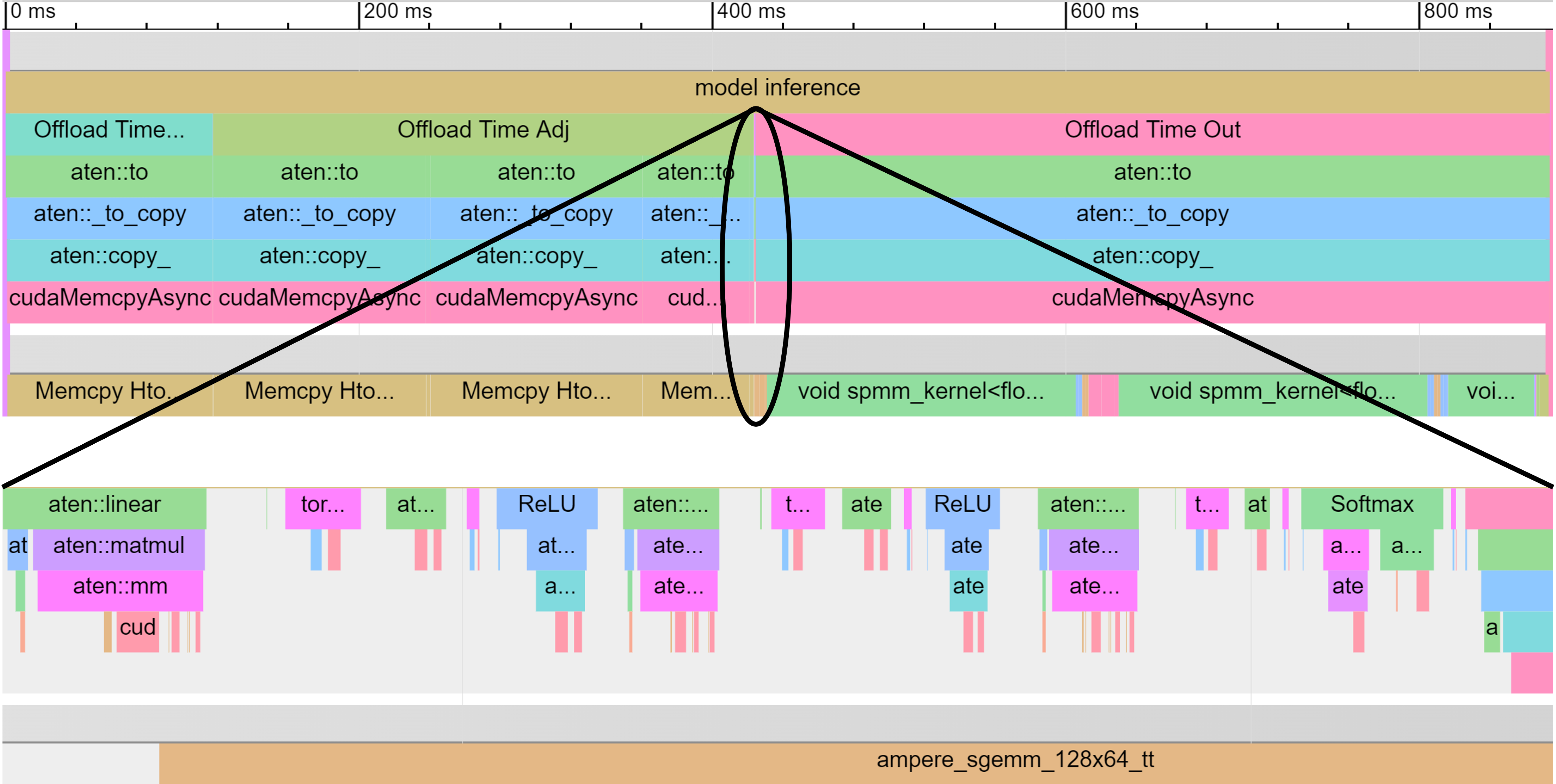}
\caption{
    GPU execution timeline for \textit{products} using embedding dimension 256. This trace shows the performance of the host CPU and GPU hardware. In the top figure, the first group of traces are for the CPU while the bottom trace is the GPU. The impact of offload time is almost 50\% for this graph when running on GPU. The sparse kernel dominates most of the remaining execution time. The expanded section shows the host offloading the GCN kernels to the GPU. Notice that after the CPU initializes a cuda kernel within the \textit{aten:linear}, the GPU \textit{ampere\_sgemm\_128x64\_tt} kernel starts (bottom row). 
}
\label{fig:gpu-exe-time}
\vspace{-1em}
\end{figure}

\begin{enumerate}
    \item Load the OGB graph in CSR format and feature matrix into CPU memory. 
    \item Construct GCN model and load weights for convolution kernels. 
    \item (GPU only) Offload GCN model weights to GPU. Do not include offload time of model in performance characterization because model parameters are assumed to be pre-loaded on GPU in production settings.
    \item Begin profiling GCN using the PyTorch profiling tool and record execution time for key kernels. 
    \item (GPU only) Offload adjacency matrix and feature matrix from CPU to GPU. 
    \item Run forward inference of graph through GCN model. See Figure \ref{fig:cpu-exe-time} for detailed execution time plot on CPU and Figure \ref{fig:gpu-exe-time} for GPU.
    \item (GPU only) Offload result of GCN from GPU to CPU. 
\end{enumerate}

Figures \ref{fig:cpu-exe-time} and \ref{fig:gpu-exe-time} show the execution timeline for full-graph GCN on CPU and GPU respectively. Both plots use the \textit{products} graph. CPU executes \textit{products} GCN in $\sim$6 seconds while GPU executes \textit{products} GCN $\sim$800 milliseconds. 
Notice that the offload features and offload adjacency matrix must occur before any GCN kernels can be started. This offload time is unavoidable and is determined by the interconnect bandwidth between the CPU and GPU. This bandwidth is typically in the 10x GBs range whereas GPU memory to GPU die is typically in the 100xGBs-TBs range. Systems with different CPU-GPU bandwidth will see this offload time scale proportionally. A performance characterization not included in this study is the impact of data movement on total energy. Data movement at the system level is expensive; future studies should reveal the impact on energy efficiency of using a GPU versus a CPU.  

All the OGB graphs were profiled on CPU for an embedding dimension sweep from 8 to 256. All graphs which fit on the GPU (all except \textit{papers}) where characterized for the same set of embedding dimensions.

\begin{figure}[t]
\centering
\includegraphics[width=.9\linewidth]{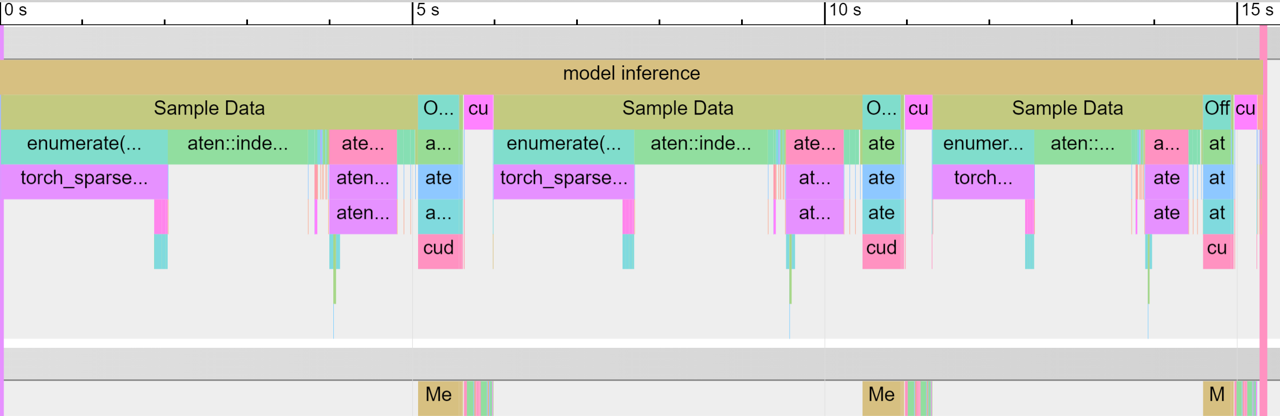}
\caption{
    GPU execution timeline for \textit{papers} using batch-wise neighborhood sampling for embedding dimension 256. The first group of traces are for the CPU while the bottom trace is the GPU. In this trace, three mini-batches are shown out of $\sim$1.7 million. The impact of neighborhood sampling dominates the total execution time (shown as \textit{Sample Data} in the CPU timeline). Offload time is the next highest impact shown in teal immediately following \textit{Sample Data}.
}
\label{fig:gpu-exe-time-sampled-batch-wise}
\vspace{-1em}
\end{figure}

\begin{figure}[t]
\centering
\includegraphics[width=.9\linewidth]{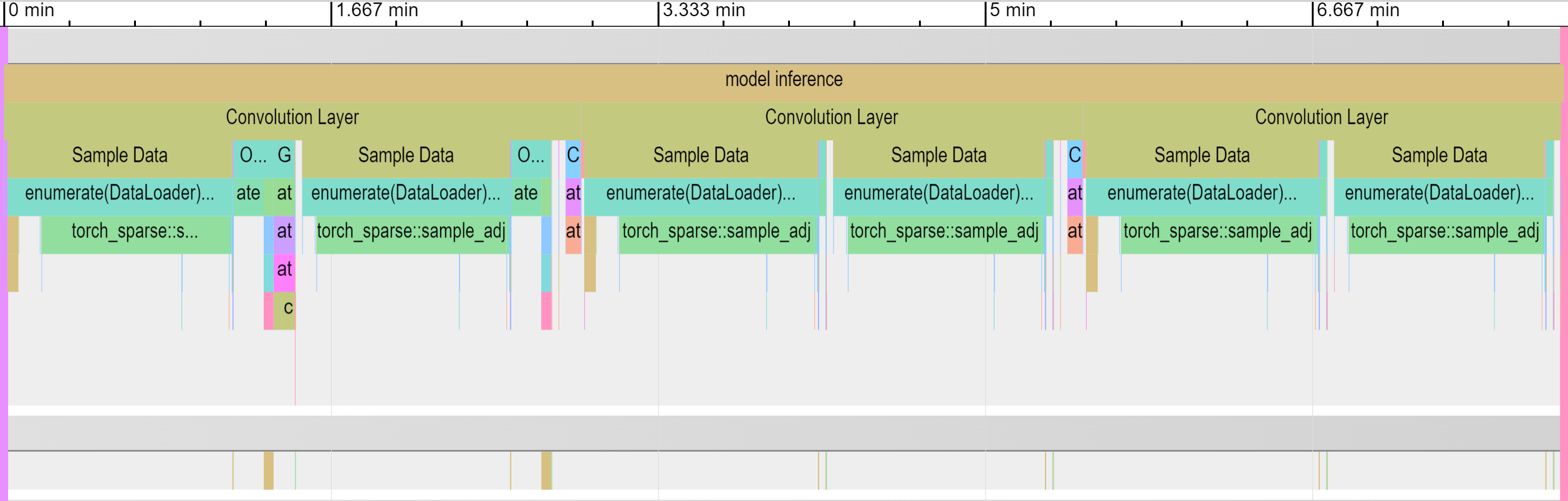}
\caption{
    GPU execution timeline for \textit{papers} using layer-wise neighborhood sampling for embedding dimension 256. The first group of traces are for the CPU while the bottom trace is the GPU. In this trace, two mini-batches are shown for each GCN layer, totalling six sampling occurrences. Only two mini-batches out of $\sim$100 per layer are shown to highlight the separation of each layers computation. 
}
\label{fig:gpu-exe-time-sampled-layer-wise}
\vspace{-1em}
\end{figure}

\subsection{Full-Neighborhood-Sampling GCN Characterization}

% batch-wise
% layer-wise pass what you need for each layer then move onto next layer

To address the challenge of graphs which do not fit in GPU memory, neighborhood sampling is required. Two methods for full-neighborhood sampling are investigated; batch-wise and layer-wise sampling. 

\subsubsection{Batch-Wise Neighborhood-Sampling}

Figure \ref{fig:gpu-exe-time-sampled-batch-wise} shows an execution timeline for three batches of \textit{papers} using batch-wise neighborhood sampling. Notice that the GPU has low utilization during each batch. This is because the CPU must first collect the data which feeds into the GPU. The GPU completes its tasks well before the CPU can prepare the next set of data, leading to severe GPU under utilization. 

Additionally, to minimize latency, the largest mini-batch size which fit on GPU was required. Experimentally, the largest batch-size per embedding dimension were discovered. Further research can optimize this batch size as the results of this work led to conservative batch sizes. 

Finding the largest batch size for a graph is important because it minimizes the data-movement between CPU and GPU. Every batch moves feature vectors from the CPU to the GPU. Only a small percent of those feature vectors are actually updated during each batch, thus the data-movement cost during sampled inference on GPU is incredibly high. The base memory footprint of \textit{papers} with embedding dimension 256 is $\sim$100GB. The additional data-movement for \textit{papers} is on the order of 100s / 1000s TBs of additional data movement! 

The largest batch size which could run on GPU for \textit{papers} with embedding dimension 256 was a batch size of 64. Using neighborhood sampling, the number of vertices for a three layer GCN expands to 1-4M vertices; meaning 99.99\% of embedding vectors were auxiliary. This led to a data-movement cost of $\sim$4GB per batch and a memory footprint of 30GB - almost reaching full memory-capacity on the A100-40GB GPU when including the model weights and other data structures. 

Furthermore, 1.7 million mini-batches (111M total vertices / 64 vertices per mini-batch) were required for \textit{papers} with embedding dimension 256, leading to an additional data-movement cost of \textbf{$\sim$2-6 Petabytes} when using full-neighborhoods. 

The data movement requirement can be approximately computed by taking the batch size and multiplying it by the average edges per vertex raised to the number of layers. In the case of \textit{papers} with embedding dimension 256, the number of embedding vectors per mini-batch equals $64*30^{3}$. 

Clearly, offloading large graphs to GPU using batch-wise full-neighborhood sampling is intractable. Fixed-neighborhood batch-wise sampling still faces the exploding neighborhood problem, which will likely cause a similar data-movement issue.

\subsubsection{Layer-Wise Neighborhood-Sampling}

Layer-wise neighborhood sampling is an alternative sampling algorithm to batch-wise sampling which trades off larger memory footprint with faster execution time. Figure \ref{fig:gpu-exe-time-sampled-layer-wise} shows an execution timeline for layer-wise sampling. Each GCN layer is broken into mini-batches, and each mini-batch requires a sampling step. In this case, there are three GCN layers where only two mini-batches per layer are shown. 

The larger memory footprint is because for each layer, the intermediate feature vectors are saved in CPU memory. The change in memory footprint depends on the input-output channel sizes for the GCN. 

Layer-wise sampling may have a larger memory footprint than batch-wise sampling because all the intermediate features must be saved on CPU before computing the next layer. For example, if a graph has an input dimension size of 128 and a hidden dimension size of 256, then the CPU memory footprint for the first GCN layer using layer-wise sampling is \#-vertices * 128 + \#-vertices * 256 whereas with batch-wise sampling, the CPU memory footprint for the first GCN layer was only \#-vertices * 128. 

%An algorithm comparison of layer-wise neighborhood sampling and batch-wise neighborhood sampling is shown in Algorithms \ref{alg:layer-wise} and \ref{algo:batch-wise}.
%\input{text/layerwise_sampling}

The case of \textit{papers} with embedding dimension 256 was examined. The largest minibatch-size for this embedding dimension was 1M vertices, leading to a maximum GPU memory footprint of $\sim$30 GB and $\sim$100 total batches. This minibatch-size was much higher than batch-wise sampling because layer-wise only expands the neighborhood of a mini-batch to 1-hop neighborhoods; reducing the impact of the exploding neighborhood problem with batch-wise sampling. 

The GPU memory requirement can be derived by taking the GCN layer with the largest convolution dimensions. The largest GCN layer in this example was the hidden layer with input dimension 256 and output dimension 256. After expanding the 1-hop neighbors for the 1M vertices in a mini-batch, the total number of vertices used in the GCN layer was $\sim$ 15M. Thus, the total memory capacity requirement on GPU was the number of vertices times the (input channel + output channel) times the data size $= 15M * (256 + 256) * 4 = \sim30$GB. 

The total data-movement can be calculated by multiplying the data-movement on and off the GPU per mini-batch by the number of mini-batches per layer by the number of layers. In the case of \textit{papers} with GCN embedding size 256, the total data movement is \textbf{$\sim$4 Terabytes}. This is three orders of magnitude lower than batch-wise neighborhood sampling, but still orders of magnitude higher than desired data movement between CPU and GPU.  
  
\begin{figure}[t]
\centering
\includegraphics[width=.9\linewidth]{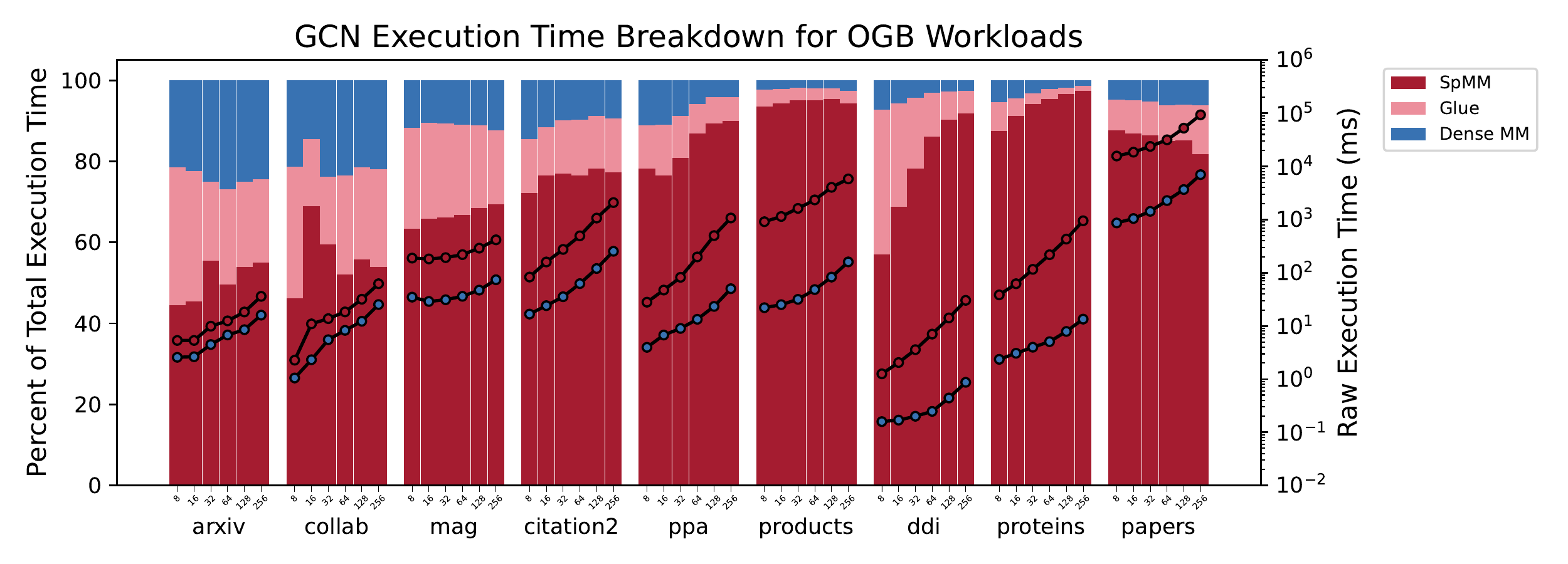}
\caption{
    Execution time breakdown for Xeon CPU.
}
\label{fig:cpu_exe_breakdown}
\vspace{-1em}
\end{figure}

\begin{figure}[t]
\centering
\includegraphics[width=.9\linewidth]{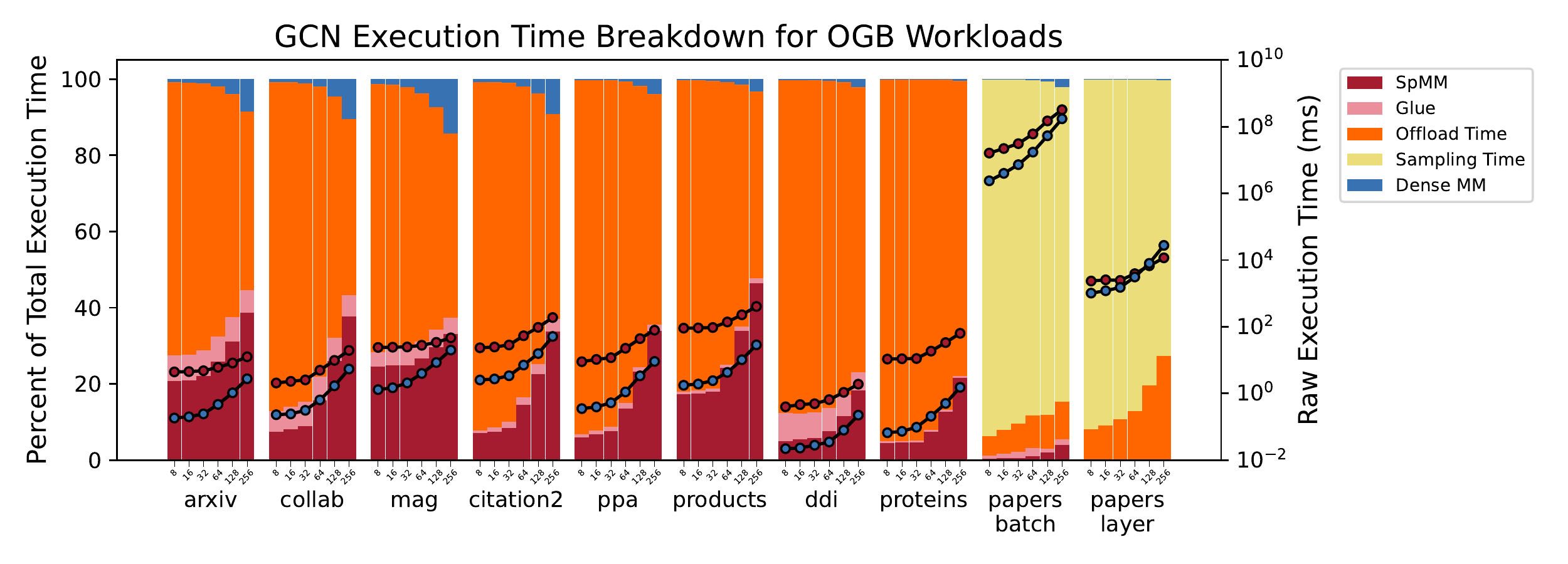}
\caption{
    Execution time breakdown for A100 GPU.
}
\label{fig:gpu_exe_breakdown}
\vspace{-1em}
\end{figure}

\section{Results}

This section explains the performance results of GCN characterized on CPU and GPU. One focus of this characterization was to breakdown the execution time of GCN into its primary components on each hardware platform; sparse matrix multiplication (SpMM), dense matrix multiplication (Dense MM), Glue Code (activation functions and kernel initialization), offload time, and sampling time.

The results shown in Figure \ref{fig:cpu_exe_breakdown} reveal several trends in CPU performance. First, SpMM is the dominant kernel in GCN, which is well understood in literature. A trend was identified that as the embedding dimension increased, the impact of the Glue Code decreased leading to higher dominance of SpMM. For example, in \textit{mag}, as the embedding dimension increased, the impact of SpMM increased while Dense MM stayed almost constant. Graphs \textit{ppa}, \textit{ddi}, and \textit{proteins} have the opposite effect; as the embedding dimension increases, the impact of the dense kernel decreases significantly. Finally, in \textit{papers}, as the embedding dimension increases, the fraction of execution time spent in SpMM decreases. 

Next, the execution time breakdown for GCN on GPU is shown in Figure \ref{fig:gpu_exe_breakdown}. All the OGB graphs except \textit{papers} fit into GPU memory. Therefore, only \textit{papers} required a sampling approach to run using GPU. The first observation was that offload time had a large impact on performance. This is well understood in literature and prior works have attempted to reduce long offload times through embedding clustering. 

Another observation was that as the embedding dimension increased, the impact of offloading to GPU decreased. Although data-transfer generally increased linearly with embedding dimension, sparse/dense multiplication increased exponentially with embedding dimension. This explains why dense and sparse had higher impact on execution time at higher embedding dimensions compared to offload time. 

Another interesting observation was that the workloads which were dominated by SpMM on CPU were not necessarily the most SpMM dominant graphs on GPU. For example, \textit{proteins} was the most SpMM dominant workload on CPU; however, \textit{products} was the most SpMM dominant on GPU. 

Finally, the \textit{papers} result shown in Figure \ref{fig:gpu_exe_breakdown} revealed the execution time breakdown for sampled GCN on GPU. From this result it is clear that sampling had the largest effect on performance. In fact, sampling and offload time constituted more than 95\% of execution time using batch-wise sampling and more than 99\% of the execution time using layer-wise sampling.

\begin{figure}[t]
\centering
\includegraphics[width=\linewidth]{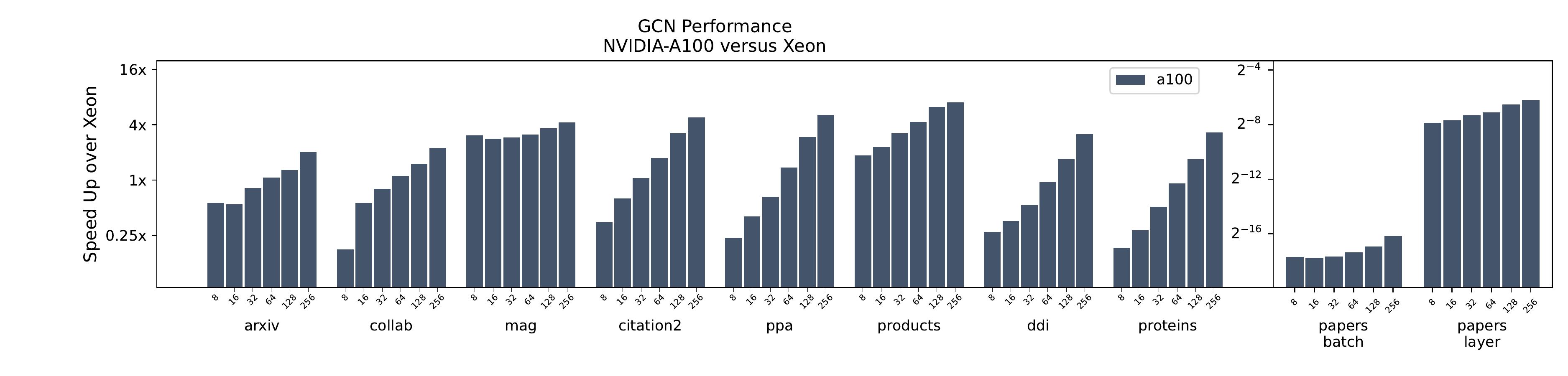}
\vspace{-2.5em}
\caption{
    Speedup of A100 GPU versus Xeon for OGB workloads. The first eight graphs fit entirely in GPU memory, while \textit{papers} required sampling, leading to much lower performance than CPU. 
}
\label{fig:cpu_versus_gpu}
\vspace{-1em}
\end{figure}

The execution times for the CPU and GPU experiments are compared in Figure \ref{fig:cpu_versus_gpu}. At higher embedding dimensions, full-graph on GPU clearly outperformed CPU; however, at low embedding dimensions, CPU had higher performance. This result is largely explained by the long offload times for GPU. As the amount of compute increased with embedding dimension, the impact of offload time decreased, as shown in Figure \ref{fig:gpu_exe_breakdown}. 

The sampling solution for \textit{papers} performs very poorly in both batch and layer sampling scenarios when compared to the CPU implementation. This was do to the performance bottleneck of CPU based sampling and data movement to GPU.

\newpage
\bibliographystyle{IEEEtran}
\bibliography{references.bib}

\end{document}